\documentclass[10pt,twocolumn,letterpaper]{article}

 \usepackage{cvpr}              

\usepackage{graphicx}
\usepackage{amsmath}
\usepackage{amssymb}
\usepackage{booktabs}
\usepackage{amsfonts}
\usepackage{bbding} 
\usepackage{color}
\usepackage{multirow}
\usepackage[accsupp]{axessibility}
\usepackage{makecell}

\usepackage[pagebackref,breaklinks,colorlinks]{hyperref}

\usepackage[capitalize]{cleveref}
\crefname{section}{Sec.}{Secs.}
\Crefname{section}{Section}{Sections}
\Crefname{table}{Table}{Tables}
\crefname{table}{Tab.}{Tabs.}

\def\cvprPaperID{13976} 
\def\confName{CVPR}
\def\confYear{2024}

\begin{document}

\title{WorldDreamer: Towards General World Models for Video Generation \\ via Predicting Masked Tokens}

\author{Xiaofeng Wang\footnotemark[1]~\textsuperscript{\rm 1}~~~Zheng Zhu\footnotemark[1]~\textsuperscript{\rm 1}\textsuperscript{\Envelope}~~~Guan Huang\footnotemark[1]~\textsuperscript{\rm 1,2}~~~Boyuan Wang\textsuperscript{\rm 1}~~~Xinze Chen\textsuperscript{\rm 1}~~~Jiwen Lu\textsuperscript{\rm 2}\\
\textsuperscript{\rm 1}GigaAI
~ ~ \textsuperscript{\rm 2}Tsinghua University
\\
\small{Project Page: \url{https://world-dreamer.github.io}}
}

\twocolumn[{%
\vspace{-1em}
\maketitle
\vspace{-1em}
\begin{center}
\centering
\resizebox{1\linewidth}{!}{
\includegraphics{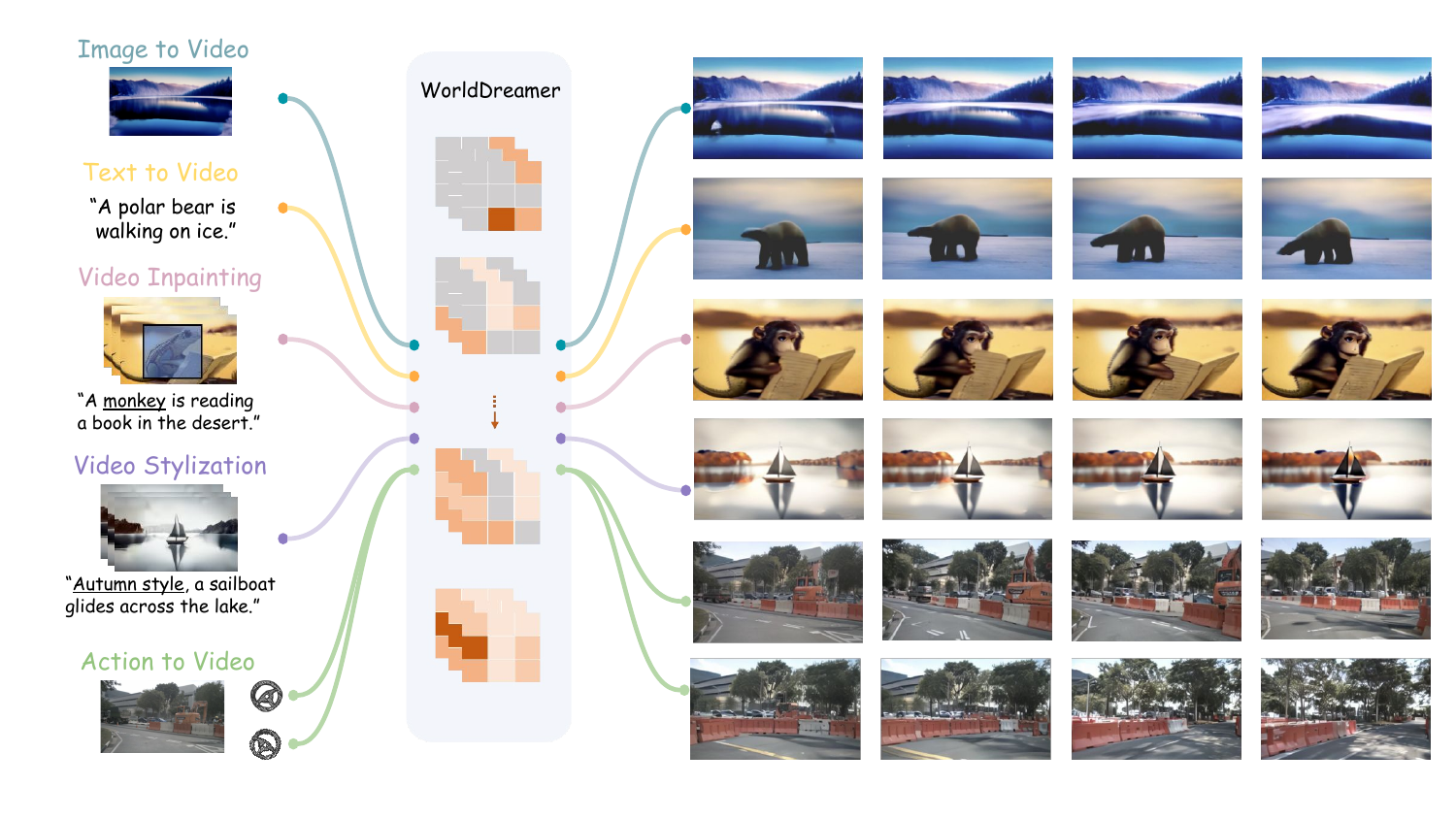}}
\captionof{figure}{\textit{WorldDreamer} demonstrates a comprehensive understanding of visual dynamics in the general world. It excels in image-to-video synthesis, text-to-video generation, video inpainting, video stylization and even action-to-video generation.}
\label{fig:main}
\end{center}}]

\renewcommand{\thefootnote}{\fnsymbol{footnote}}
\footnotetext[1]{These authors contributed equally to this work.
\textsuperscript{\Envelope}Corresponding author: Zheng Zhu, zhengzhu@ieee.org}

\begin{abstract}
\vspace{-0.3em}

World models play a crucial role in understanding and predicting the dynamics of the world, which is essential for video generation. However, existing world models are confined to specific scenarios such as gaming or driving, limiting their ability to capture the complexity of general world dynamic environments. Therefore, we introduce \textit{WorldDreamer}, a pioneering world model to foster a comprehensive comprehension of general world physics and motions, which significantly enhances the capabilities of video generation.
Drawing inspiration from the success of large language models, \textit{WorldDreamer} frames world modeling as an unsupervised visual sequence modeling challenge. This is achieved by mapping visual inputs to discrete tokens and predicting the masked ones. During this process, we incorporate multi-modal prompts to facilitate interaction within the world model.
Our experiments show that \textit{WorldDreamer} excels in generating videos across different scenarios, including natural scenes and driving environments. \textit{WorldDreamer} showcases versatility in executing tasks such as text-to-video conversion, image-to-video synthesis, and video editing. These results underscore \textit{WorldDreamer}'s effectiveness in capturing dynamic elements within diverse general world environments.

\end{abstract}

\section{Introduction}

The next significant leap in artificial intelligence is expected to come from systems that possess a profound understanding of the dynamic visual world. At the core of this advancement are world models, crucial for comprehending and predicting the dynamic nature of our world. World models hold great promise for learning motion and physics in the general world, which is essential for video generation. 

The early exploration of world models \cite{worldmodel} primarily focus on gaming scenarios, which proposes a generative neural network model capable of learning compressed representations of spatial and temporal dynamics within game environments. Subsequent research in the Dreamer series \cite{dreamv1,dreamv2,dreamv3} further validated the efficacy of world models across diverse gaming scenarios. Considering its structured nature and paramount importance, autonomous driving has become a forefront domain for the practical application of world models. Various approaches \cite{drivedreamer,hu2023gaia,adriver,drivewm} are introduced to explore the efficacy of world models in autonomous driving scenarios. Furthermore, DayDreamer \cite{wm4} has extended the application of world models to encompass real-world robotic environments, However, current world models are predominantly confined to gaming, robotics, and autonomous driving, lacking the capability to capture the motion and physics of the general world. Besides, relevant research in world models mainly relies on Recurrent Neural Networks (RNNs) \cite{wm1,wm2,wm3,wm4,wm5,dreamv1,dreamv2,dreamv3} and diffusion-based methods \cite{drivedreamer,drivewm,adriver} to model visual dynamics. While these approaches have yielded some success in video generation, they encounter challenges in effectively capturing the motion and physics in general world scenes.

In this paper, we introduce \textit{WorldDreamer}, which pioneers the construction of general world models for video generation. Drawing inspiration from the successes of large language models (LLMs) \cite{gpt1,gpt2,gpt3,bert1}, we predict the masked visual tokens to effectively model the intricate dynamics of motion and physics embedded in visual signals. Specifically, \textit{WorldDreamer} involves encoding images into discrete tokens using VQGAN \cite{vqgan}. We then randomly mask a portion of these tokens and utilize the unmasked tokens to predict the masked ones, a process integral to capturing the underlying motion and physics in visual data. \textit{WorldDreamer} is constructed on the Transformer architecture \cite{vaswani2017attention}. Regarding the spatial-temporal priority inherent in video signals, we propose the Spatial Temporal Patchwise Transformer (STPT), which enables attention to focus on localized patches within a temporal-spatial window, facilitating the learning of visual signal dynamics and accelerating the convergence of the training process. Additionally, \textit{WorldDreamer} integrates language and action signals through cross-attention, to construct multi-modal prompts for interaction within world model. Notably, compared to diffusion-based methods, \textit{WorldDreamer} capitalizes on the reuse of LLM infrastructure and benefits from optimizations developed over years for LLMs, including model scaling learning recipes. Besides, \textit{WorldDreamer} exhibits a remarkable speed advantage, parallel decoding videos with just a few iterations, which is $\sim$3$\times$ faster than diffusion-based methods \cite{blattmann2023stable,modelscope,videocraft}. Therefore, \textit{WorldDreamer} holds great promise for constructing a general world model from visual signals.

The main contributions of this paper can be summarized as follows: (1) We introduce \textit{WorldDreamer}, the first general world model for video generation, which learns general world motion and physics. (2) We propose the Spatial Temporal Patchwise Transformer (STPT), which enhances the focus of attention on localized patches within a temporal-spatial window. This facilitates easier learning of visual signal dynamics and expedites the training process. (3) We conduct extensive experiments to verify that \textit{WorldDreamer} excels in generating videos across different scenarios, including natural scenes and driving environments. \textit{WorldDreamer} showcases versatility in executing tasks such as text-to-video conversion, image-to-video synthesis, video editing, and action-to-video generation (see Fig.~\ref{fig:main}).

\begin{figure*}[ht]
\centering
\resizebox{0.92\linewidth}{!}{
\includegraphics{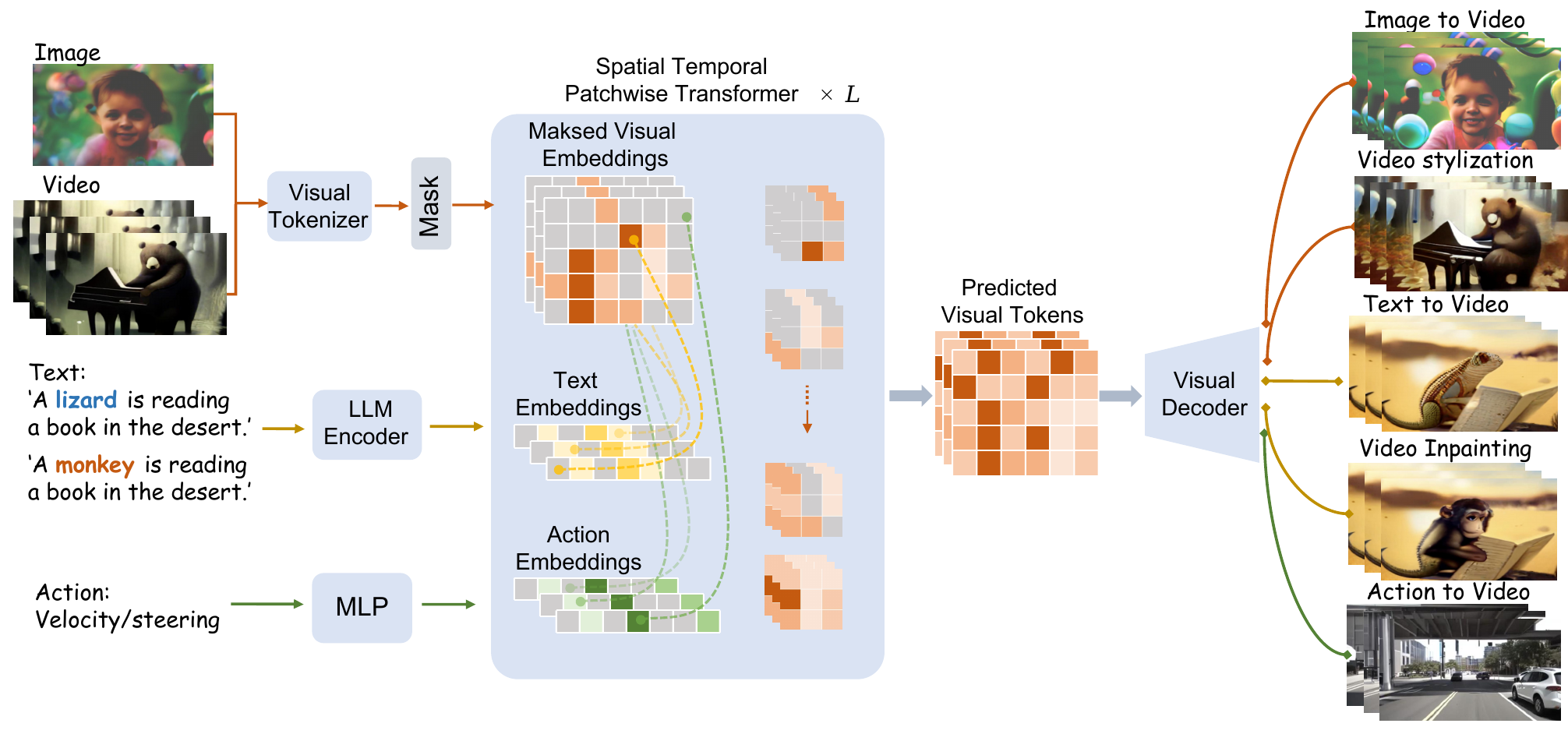}}
\caption{Overall framework of \textit{WorldDreamer}. \textit{WorldDreamer} first converts images and videos into visual tokens, followed by a token-masking operation. Text and action inputs are encoded separately into embeddings, acting as multimodal prompts. Subsequently, STPT predicts the masked visual tokens, which are processed by visual decoders to enable video generation and editing in various scenarios.}
\label{fig:wd}
\end{figure*}

\section{Related Work}
\subsection{Video Generation}
Currently, state-of-the-art video generation models are primarily classified into two categories: Transformer-based methods and diffusion-based methods.

\textbf{Transformer-based methods.} 
The Transformer-based video generation methods are derived from the general family of LLMs \cite{gpt1,gpt2,gpt3,bert1}. Typically, these methods employ autoregressive prediction of the next token or parallel decoding of masked tokens to generate videos. Drawing inspiration from image generation techniques \cite{dalle,cogview,imagegpt,parti}, VideoGPT \cite{yan2021videogpt} integrates VQVAE \cite{vqvae} with Transformer-based token prediction, enabling it to autoregressively predict visual tokens for video generation.
Furthermore, GAIA-1 \cite{hu2023gaia} integrates various modalities, including text descriptions, images, and driving actions, resulting in the generation of autonomous driving scenario videos. Unlike these autoregressive methods, some Transformer-based approaches \cite{hong2022cogvideo,villegas2022phenaki}, draw inspiration from \cite{magvit,maskgit,chang2023muse,ding2022cogview2}, accelerating video generation through parallel decoding. In addition to these methods, VideoPoet \cite{kondratyuk2023videopoet} adopts video tokenizer \cite{magvit2} and generates exceptionally high-quality videos based on parallel decoding. The incorporation of Transformer models into video language models showcases their formidable zero-shot capability in handling various tasks during pretraining. Therefore, employing Transformer-based mask image models as the foundation for general world models emerges as a promising avenue.

\textbf{Diffusion based methods.} Compared to Transformer-based models, there has been extensive research employing diffusion-based models for video generation. 
VideoLDM \cite{blattmann2023align} introduces a temporal dimension to the latent space of the 2D diffusion model and fine-tuned it using videos, effectively transforming the image generator into a video generator and enabling high-resolution video synthesis. Similarly, LVDM \cite{he2022latent} explores lightweight video diffusion models, making use of a low-dimensional 3D latent space. Make-A-Video \cite{singer2022make} also employs a pre-trained text-to-image model, eliminating the need for large-scale video training. Moreover, in the Imagen Video \cite{ho2022imagen}, a cascading video diffusion model is built upon the pretrained 2D diffusion model \cite{ho2022imagen}. DiffT \cite{diffit} and W.A.L.T \cite{gupta2023photorealistic} improve the video generation by utilizing a Transformer-based Diffusion network. Recently, Emu Video \cite{emuvideo} and PixelDance \cite{pixeldance} propose a two-step factorization approach for text-to-video generation, wherein the process is initially decomposed into text-to-image conversion, followed by image-to-video synthesis. This methodology capitalizes on the effectiveness of contemporary text-to-image models, strategically directing the focus of the video diffusion model training toward the learning of motion dynamics. However, diffusion-based methods have difficulty integrating multiple modalities within a single model. Furthermore, these diffusion-based approaches struggle to produce results that accurately capture dynamics and motion.

\subsection{World Models}
World models play a pivotal role in comprehending and predicting the dynamic nature of our environment, holding immense potential for acquiring insights into motion and physics on a global scale. Initially, the exploration of world model \cite{worldmodel} focuses primarily on gaming scenarios, presenting a generative neural network model capable of learning condensed representations of spatial and temporal dynamics within game environments. Subsequent research within the Dreamer series \cite{dreamv1,dreamv2,dreamv3} affirmed the effectiveness of world models across a diverse array of gaming scenarios. Given its structured nature and critical significance, the domain of autonomous driving has emerged as a forefront application area for world models. Numerous approaches \cite{drivedreamer,hu2023gaia,drivewm,adriver} have been introduced to assess the efficacy of world models in autonomous driving scenarios. Additionally, DayDreamer \cite{wm4} has expanded the scope of world models to encompass real-world robotic environments. However, it is noteworthy that current world models primarily operate within the realms of gaming, robotics, and autonomous driving, lacking the capability to comprehensively capture the motion and physics of the general world.

\section{WorldDreamer}
\subsection{Overall Framework}
The overall framework of \textit{WorldDreamer} is depicted in Fig.~\ref{fig:wd}. The initial phase involves encoding visual signals (\ie, images and videos) into discrete tokens using a visual tokenizer. These tokens undergo a carefully devised masking strategy before being processed by STPT. Meanwhile, textual and action signals are separately encoded into embeddings, which serve as multimodal prompts. STPT engages in the pivotal task of predicting the masked visual tokens, which are then decoded by visual decoders, facilitating video generation and editing in multiple contexts.

To train \textit{WorldDreamer}, we construct triplets of \textit{Visual-Text-Action} data, where the training supervision solely involves predicting masked visual tokens without any additional supervision signals. \textit{WorldDreamer} also supports training without text or action data, which not only reduces the difficulty of data collection but also enables \textit{WorldDreamer} to learn unconditional or single-condition video generation. At inference time, \textit{WorldDreamer} can accomplish various video generation and video editing tasks: (1) For image-to-video, only a single image input is needed, considering the remaining frames as masked. \textit{WorldDreamer} can also predict the future frames based on both single image condition and text condition. (2) For video stylization, a video segment can be input, with a random masking of certain pixels. \textit{WorldDreamer} can alter the video style, such as creating an autumn-themed effect, based on both the input language. (3) For text-to-video, providing language input allows \textit{WorldDreamer} to predict the corresponding video, assuming that all visual tokens are masked. (4) For video inpainting, a video segment can be input, with a manually masked region of interest. \textit{WorldDreamer} can fill in the masked portion based on the input language and unmasked visual signals. (5) For action-to-video, inputting the initial frame of a driving scene along with future driving commands allows \textit{WorldDreamer} to predict future frames.

The subsequent subsections elaborate on the model architecture and the masking strategy. 

\subsection{Model Architecture}

\noindent
\textbf{Preliminery} \textit{WorldDreamer} utilizes VQGAN \cite{esser2021taming} to tokenize visual signals:
\begin{equation}
    V = \mathcal{F}_v(I),
\end{equation}
where $I\in\mathcal{R}^{N\times H\times W\times 3}$ are $N$ frames of visual inputs. VQGAN $\mathcal{F}_v$ downsamples the resolution by 16$\times$, which produces visual tokens $T_{\text{V}}\in\mathcal{R}^{N\times h \times w}$ ($h=\frac{H}{4}, w=\frac{W}{4}$). The VQGAN has a vocabulary size of 8192 and is trained with billions of images \cite{schuhmann2022laion}. For text inputs, we employ the pretrained T5 \cite{raffel2020exploring} to map them into high-dimensional embeddings $E_{\text{T}}\in\mathcal{R}^{K\times C_{\text{T}}}$, where $K$ is the sequence length and $C_{\text{T}}$ is the embedding channel. To be compatible with the feature learning in STPT, the text embeddings are repeated for $N$ frames, and the embedding channel is mapped into $C_{\text{V}}$.  Furthermore, Multi Layer Perception (MLP) is utilized to encode action inputs, which generates action embeddings $E_{\text{A}}\in\mathcal{R^{N\times C_{\text{V}}}}$. The text embeddings and action embeddings are concatenated, producing the multimodal prompt embeddings $E_{\text{M}}\in\mathcal{R}^{N\times (K+1) \times C_{\text{V}}}$. Note that either text or action embedding can be empty, enabling unconditional learning. 

During training, and the optimizing objective is to predict the masked visual token conditioned on unmasked tokens and multimodal prompts:
\begin{equation}
    \mathcal{L}_{\text{WorldDreamer}} = -\log p(\hat{T}_{\text{V}}| \widetilde{T}_{\text{V}}, E_{\text{M}}),
\end{equation}
where $\hat{T}_{\text{V}}$ are masked visual tokens, and $\widetilde{T}_{\text{V}}$ are unmasked visual tokens.

\begin{figure}[t]
\centering
\resizebox{1\linewidth}{!}{
\includegraphics{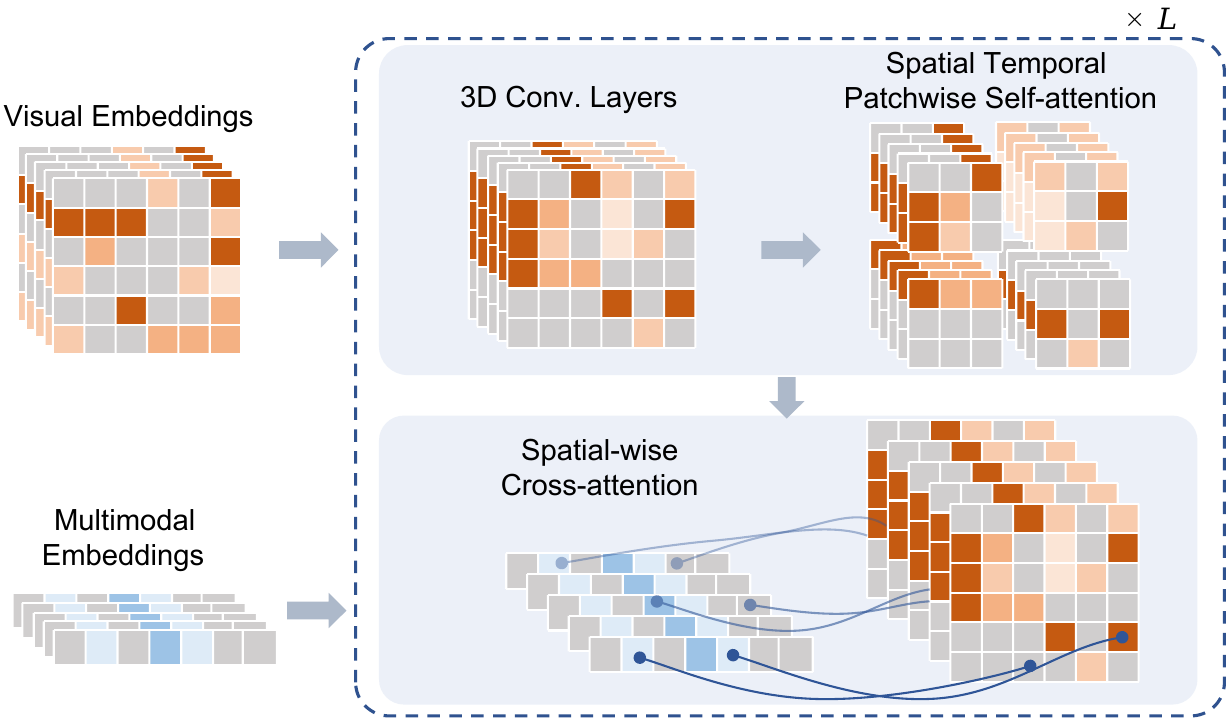}}
\caption{Overall architecture of STPT. STPT first utilizes 3D convolution to aggregate visual embeddings. Then these embeddings are partitioned into several patches for spatial-temporal patchwise self-attention.In the following, spatial-wise cross attention is applied to facilitate feature interaction between visual embeddings and multimodal embeddings.}
\label{fig:stpt}
\end{figure}

\begin{figure*}[t]
\centering
\resizebox{0.95\linewidth}{!}{
\includegraphics{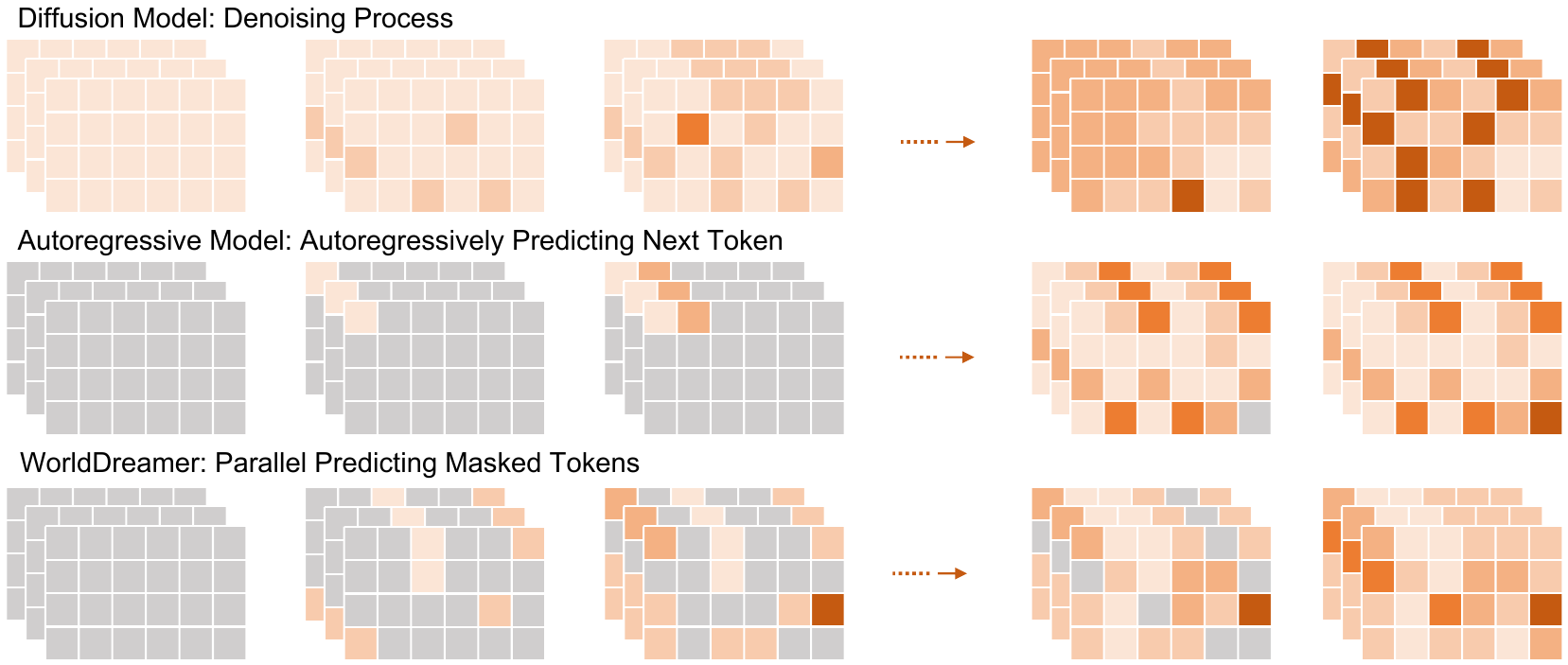}}
\caption{Comparison between the inference schedule of diffusion-based methods, autoregressive methods and \textit{WorldDreamer}. Diffusion-based methods usually require  $\sim$30 steps to reduce noise, and autoregressive methods need $\sim$200 steps to iteratively predict the next token. In contrast, \textit{WorldDreamer} parallel predicts masked tokens, achieving video generation in about 10 steps.}
\label{fig:inf}
\vspace{-1em}
\end{figure*}

\noindent
\textbf{STPT}
STPT leverages the foundation of U-ViT \cite{hoogeboom2023simple} while strategically enhancing its architecture to better capture the intricacies of spatial-temporal dynamics in video data. Specifically, STPT confines the attention mechanism within spatial-temporal patches. Additionally, to seamlessly incorporate multimodal information, spatial-wise cross-attention is employed to integrate multimodal embeddings. For input tokens $\hat{T}_{\text{V}}$, STPT transforms them into visual embeddings $E_{\text{V}}\in\mathcal{R}^{N\times h \times w \times C_{\text{V}}}$ by referencing a learnable codebook. The size of this codebook is set to 8193, exceeding the codebook size of VQGAN by 1, which enables compatibility with masked tokens. In each layer of STPT, as illustrated in Fig.~\ref{fig:stpt}, the visual embeddings are first processed through a 3D convolutional network. Then these embeddings are spatially partitioned into several patches $E_{\text{P}}\in\mathcal{R}^{N\times h/s \times w/s \times C_{\text{V}}}$, where we empirically set patch stride $s$ as 2. Subsequently, each patch embeddings are flattened for spatial-temporal patchwise self-attention:
\begin{equation}
    E_{\text{P}} = \mathcal{G}_s^{-1}(\mathcal{F}_{\text{st}}(\mathcal{G}_s(E_{\text{P}}))),
\end{equation}
where $\mathcal{G}_s$ is the flatten operation that maps the embedding dimension to $\mathcal{R}^{Nhw/s^2 \times C_{\text{V}}}$, and $\mathcal{G}_s^{-1}$ is the reverse operation. $\mathcal{F}_{\text{s}}$ is the standard self-attention. These patches are then concatenated and reshaped back to their original dimensions. In the following, the spatial-wise cross attention is applied, which facilitates feature interaction between visual embeddings and multimodal embeddings:
\begin{equation}
    E_{\text{V}} = \mathcal{F}_{\text{c}}(E_{\text{v}}, E_{\text{M}}),
\end{equation}
where $\mathcal{F}_{\text{c}}$ is the cross-attention operation that regards the frame number as batch size.
After being processed through $L$ layers of STPT, the feature dimensionality of $E_{\text{V}}$ is mapped to the codebook size of VQGAN. This enables the utilization of softmax to calculate the probability of each token, facilitating the prediction of masked visual tokens. Finally, cross-entropy loss is employed to optimize the proposed STPT:
\begin{equation}
    \mathcal{L}_{\text{ce}}(\hat{T}_{\text{V}}, \mathcal{P}_{\text{STPT}}(\widetilde{T}_{\text{V}},E_{\text{M}})),
\end{equation}
where $\mathcal{P}_{\text{STPT}}(\widetilde{T}_{\text{V}},E_{\text{M}})$ are visual token probabilities predicted by the STPT.

Notably, the proposed STPT can be trained jointly with videos and images. For image inputs, we simply replace the attention weight of $\mathcal{F}_{\text{s}}$ as a diagonal matrix \cite{gupta2023photorealistic}. Simultaneous training on both video and image datasets offers a substantial augmentation of the training samples, enabling more efficient utilization of extensive image datasets. Besides, the joint training strategy has significantly enhanced \textit{WorldDreamer}'s capability to comprehend temporal and spatial aspects within visual signals.

\subsection{Mask Strategy}
Mask strategy is crucial for training \textit{WorldDreamer}, following \cite{chang2023muse}, we train \textit{WorldDreamer} utilizing a dynamic masking rate based on cosine scheduling. Specifically, we sample a random mask rate $r\in[0,1]$ in each iteration, and totally $\frac{2hw}{\pi}(1-r^2)^{\frac{-1}{2}}$ tokens are masked in each frame.  Note that we employ the same token mask across different frames. This decision is grounded in the similarity of visual signals between adjacent frames. Using different token masks could potentially lead to information leakage during the learning process. 
In comparison to an autoregressive mask scheduler, the dynamic mask schedule employed in our approach is crucial for parallel sampling at inference time, which enables the prediction of multiple output tokens in a single forward pass. This strategy capitalizes on the assumption of a Markovian property, where many tokens become conditionally independent given other tokens \cite{chang2023muse}. The inference process also follows a cosine mask schedule, selecting a fixed fraction of the highest-confidence masked tokens for prediction at each step. Subsequently, these tokens are unmasked for the remaining steps, effectively reducing the set of masked tokens. As shown in Fig.~\ref{fig:inf}, diffusion-based methods usually require  $\sim$30 steps to reduce noise, and autoregressive methods need $\sim$200 steps to iteratively predict the next token. In contrast, \textit{WorldDreamer},  parallel predicts masked tokens in about 10 steps, presenting a $3\times\sim 20\times$ acceleration compared to diffusion-based or autoregressive methods.

\section{Experiment}
\subsection{Datasets}
We employ a diverse set of images and videos to train \textit{WorldDreamer}, enhancing its understanding of visual dynamics. The specific data utilized in this training includes:

\textbf{Deduplicated LAION-2B \cite{laurenccon2023obelisc}} The original LAION dataset \cite{schuhmann2022laion} presented challenges such as data duplication and discrepancies between textual descriptions and accompanying images. We follow \cite{patil2024amused} to address these issues. Specifically, we opted to utilize the deduplicated LAION-2B dataset \cite{laurenccon2023obelisc} for training \textit{WorldDreamer}. This refined dataset excludes images with a watermark probability exceeding 50\% or an NSFW probability surpassing 45\%. The deduplicated LAION dataset was made available by \cite{schuhmann2022laion}, following the methodology introduced in \cite{webster2023duplication}.

\textbf{WebVid-10M \cite{bain2021frozen}} WebVid-10M comprises approximately 10 million short videos, each lasting an average of 18 seconds and primarily presented in the resolution of $336 \times 596$. Each video is paired with associated text correlated with the visual content. A challenge posed by WebVid-10M is the presence of watermarks on all videos, resulting in the watermark being visible in all generated video content. Therefore, we opted to further refine \textit{WorldDreamer} leveraging high-quality self-collected video-text pairs.

\textbf{Self-collected video-text pairs}
We obtain publicly available video data from the internet and apply the procedure detailed in \cite{blattmann2023stable} to preprocess the obtained videos. Specifically, we use PySceneDetect\cite{castellano2020pyscenedetect} to detect the moments of scene switching and obtain video clips of a single continuous scene. Then, we filtered out clips with slow motion by calculating optical flow. Consequently, 500K high-quality video clips are obtained for training. For video caption, we extract the 10th, 50th, and 90th percentile frames of the video as keyframes. These key frames are processed by Gemini \cite{team2023gemini} to generate captions for each keyframe. Additionally, Gemini is instructed to aggregate these individual image captions into an overall caption for the entire video. Regarding that highly descriptive captions enhance the training of generative models \cite{betker2023improving}, we prompt Gemini to generate captions with as much detail as possible. The detailed captions allow \textit{WorldDreamer} to learn more fine-grained text-visual correspondence.

\textbf{NuScenes \cite{nuscenes} }
NuScenes is a popular dataset for autonomous driving, which comprises a total of 700 training videos and 150 validation videos. Each video includes approximately 20 seconds at a frame rate of 12Hz. \textit{WorldDreamer} utilizes the front-view videos in the training set, with a frame interval of 6 frames. In total, there are approximately 28K driving scene videos for training. For video caption, we prompt Gemini to generate a detailed description of each frame, including weather, time of the day, road structure, and important traffic elements. Then Gemini is instructed to aggregate these image captions into an overall caption for each video. Furthermore, we extract the yaw angle and velocity of the ego-car as the action metadata.

\subsection{Implementation Details}
\textbf{Train details}
\textit{WorldDreamer} is first trained on a combination of WebVid and LAION datasets. For WebVid videos, we extract 16 frames as a training sample. For the LAION dataset, 16 independent images are selected as a training sample. Each sample is resized and cropped to an input resolution of $256\times256$. \textit{WorldDreamer} is trained over 2M iterations with a batch size of 64. The training process involves the optimization with AdamW and a learning rate of $5\times 10^{-5}$, weight decay 0.01. To enhance the training and extend the data scope, \textit{WorldDreamer} is further finetuned on self-collected datasets and nuScenes data, where all (1B) parameters of STPT can be trained. During the finetuning stage, the input resolution is $192\times 320$, and each sample has 24 frames. \textit{WorldDreamer} is finetuned over 20K iterations with a batch size of 32, and the learning rate is $1\times 10^{-5}$.

\textbf{Inference details}
At inference time, Classifier-Free Guidance (CFG)\cite{ho2022classifier} is utilized to enhance the generation quality. Specifically, we randomly eliminate multimodal embeddings for 10\% of training samples. During inference, we calculate a conditional logit $c$ and an unconditional logit $u$ for each masked token. The final logits $g$ are then derived by adjusting away from the unconditional logits by a factor of $\beta$, referred to as the guidance scale:
\begin{equation}
    g = (1+\beta)c-\beta u.
\end{equation}
For the predicted visual tokens, we employ the pretrained VQGAN decoder to directly output the video. Notably, \textit{WorldDreamer} can generate a video consisting of 24 frames at a resolution of $192\times 320$, which takes only 3 seconds on a single A800.

\begin{figure*}[ht]
\centering
\resizebox{1\linewidth}{!}{
\includegraphics{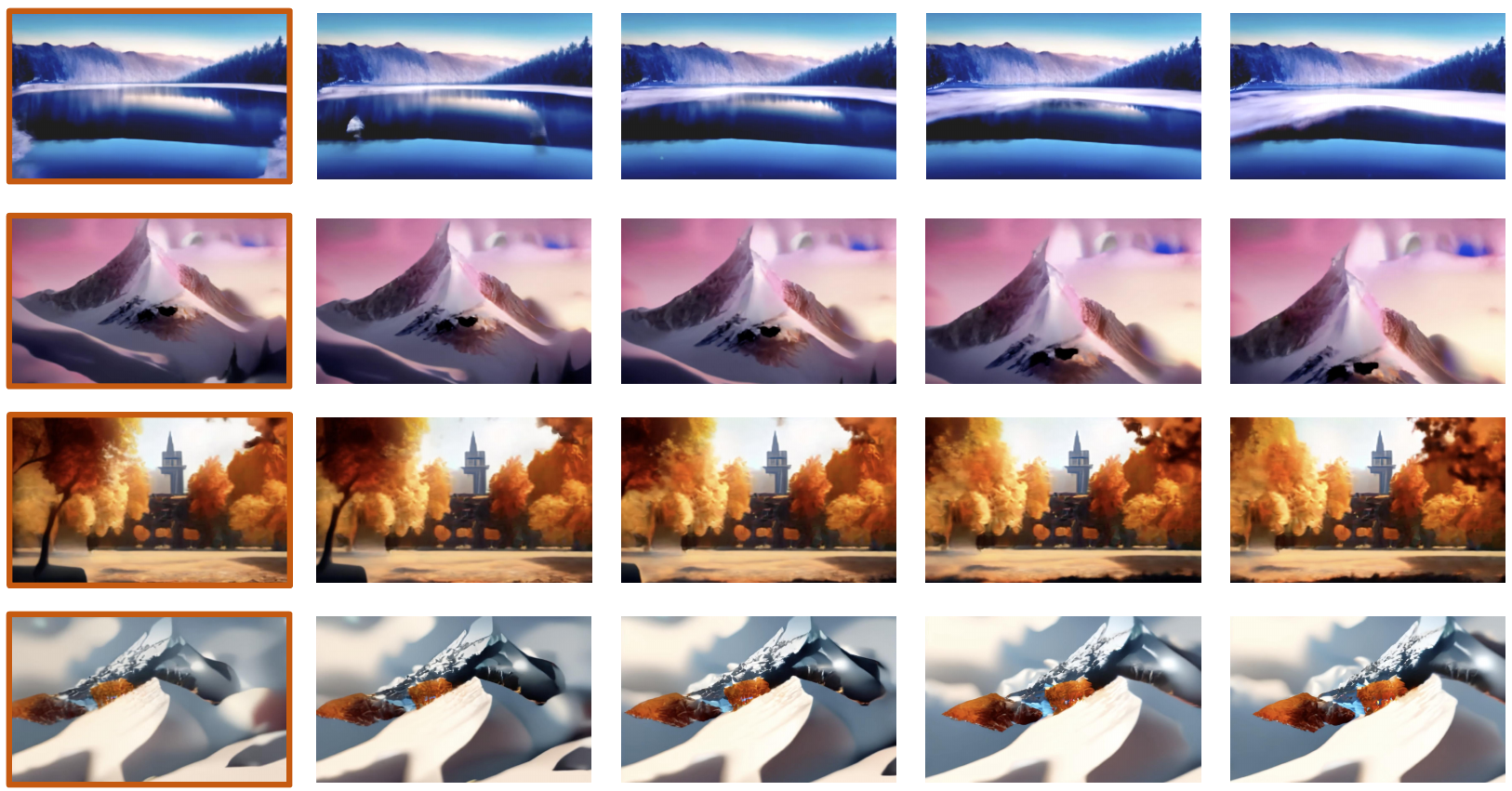}}
\caption{\textit{WorldDreamer} excels in producing high-fidelity image-to-video generation across various scenarios.}
\label{fig:i2v}
\end{figure*}

\begin{figure*}[ht]
\centering
\resizebox{1\linewidth}{!}{
\includegraphics{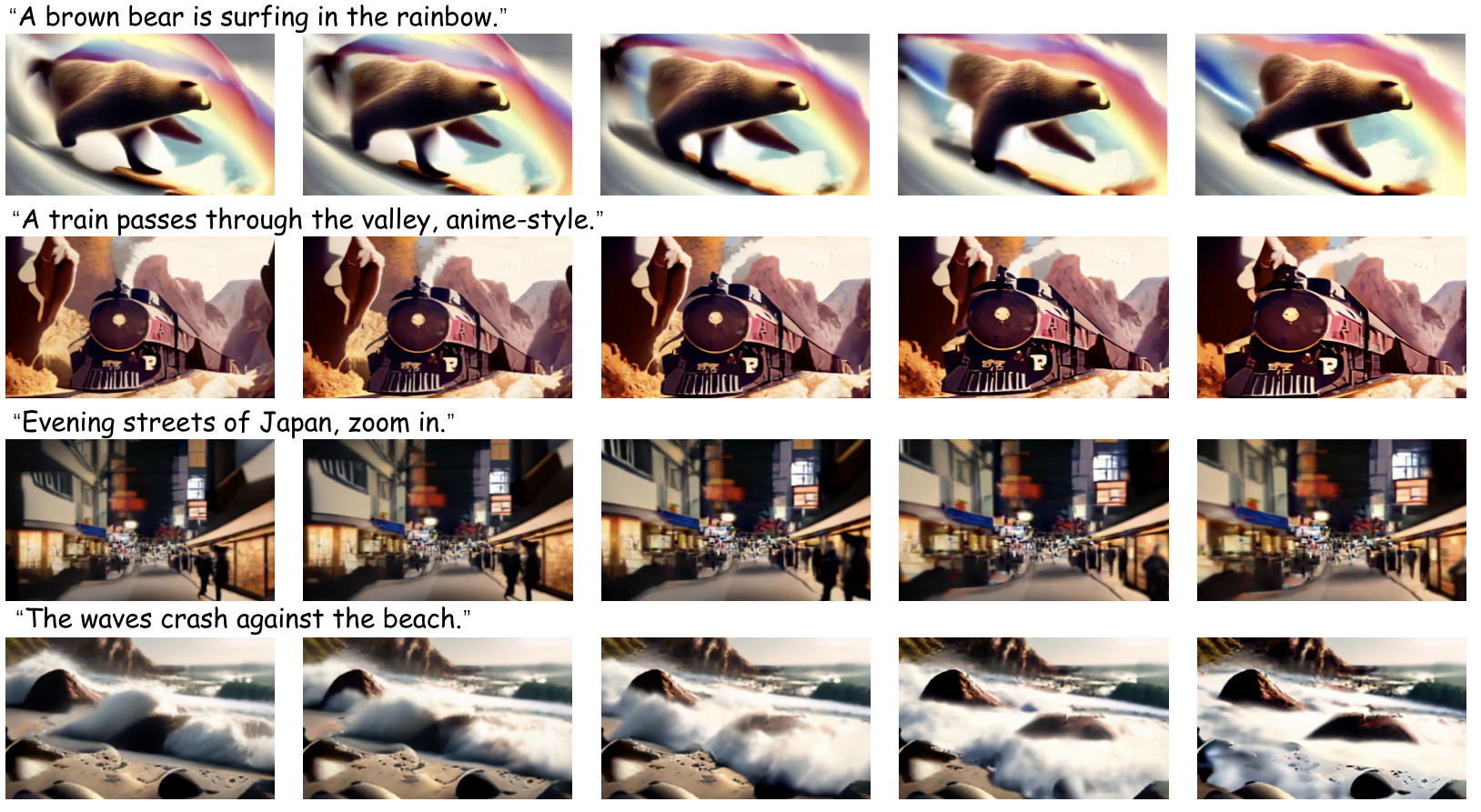}}
\caption{\textit{WorldDreamer} demonstrates proficiency in generating videos from text in diverse stylistic paradigms.}
\label{fig:t2v}
\end{figure*}

\begin{figure*}[ht]
\centering
\resizebox{1\linewidth}{!}{
\includegraphics{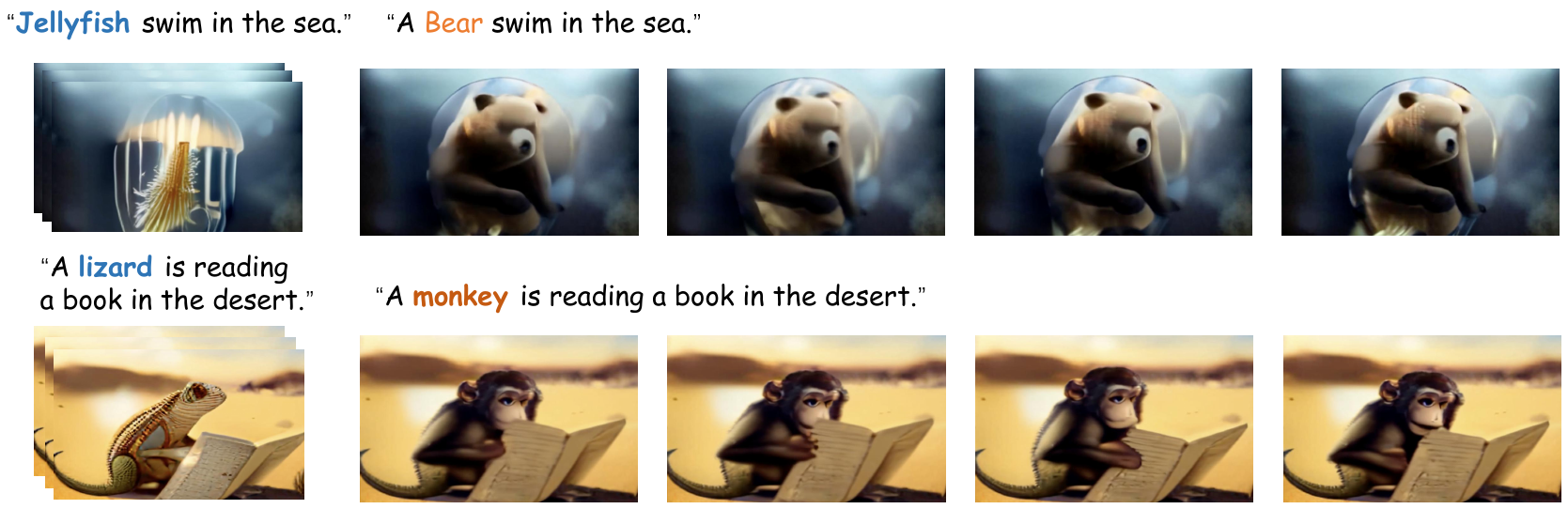}}
\caption{\textit{WorldDreamer} possesses an exceptional ability to achieve high-quality video inpainting.}
\label{fig:inpaint}
\vspace{1em}
\end{figure*}

\begin{figure*}[ht]
\centering
\resizebox{1\linewidth}{!}{
\includegraphics{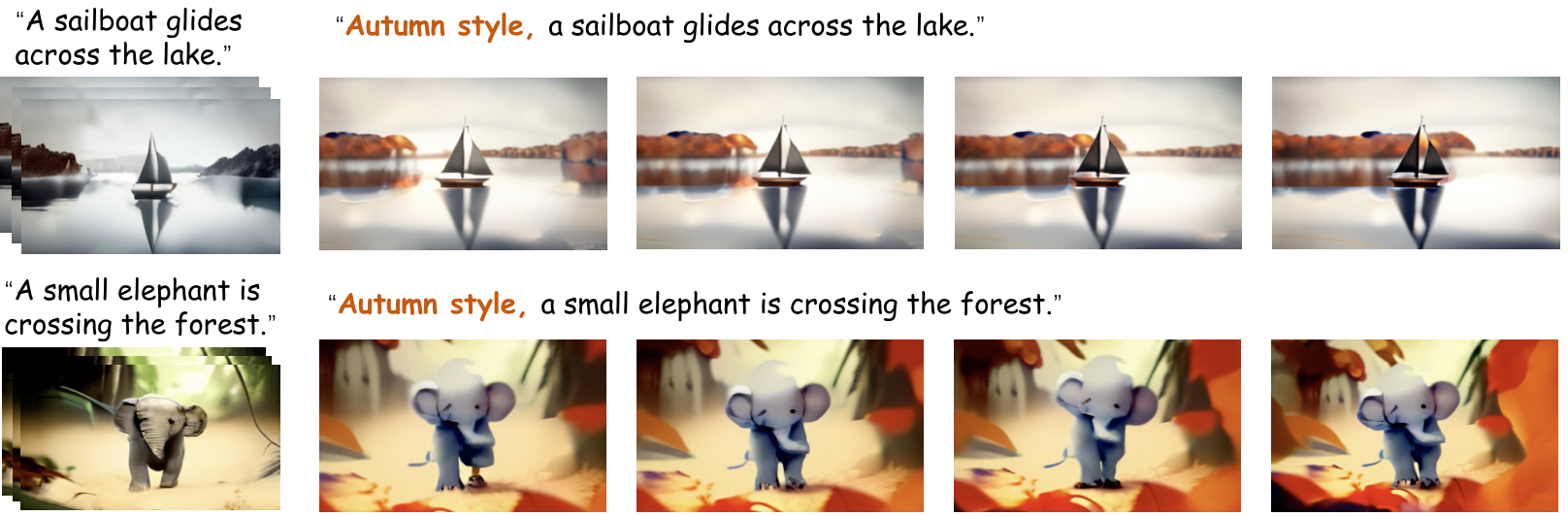}}
\caption{\textit{WorldDreamer} excels in delivering high-quality video stylization capabilities.}
\label{fig:style}
\end{figure*}

\begin{figure*}[h]
\centering
\resizebox{1\linewidth}{!}{
\includegraphics{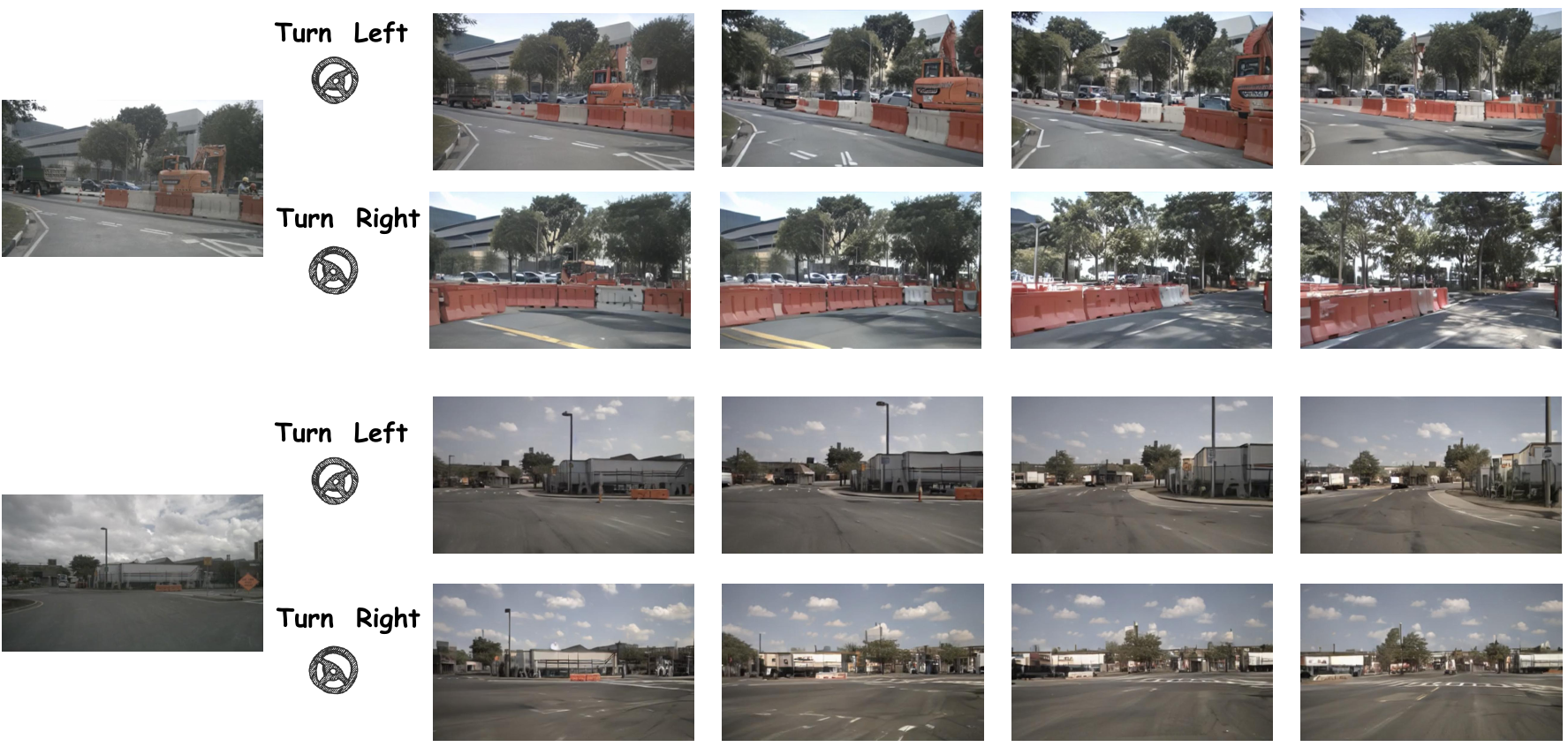}}
\caption{\textit{WorldDreamer} excels in realizing the ability to generate videos based on actions in the context of autonomous driving.}
\label{fig:ad}
\end{figure*}

\subsection{Visualizations}
We have conducted comprehensive visual experiments to demonstrate that \textit{WorldDreamer} has acquired a profound understanding of the general visual dynamics of the general world. Through detailed visualizations and results, we present compelling evidence showcasing \textit{Worlddreamer}'s ability to achieve video generation and video editing across diverse scenarios.

\textbf{Image to Video}
\textit{WorldDreamer} excels in high-fidelity image-to-video generation across various scenarios. As illustrated in Fig.~\ref{fig:i2v}, based on the initial image input, \textit{Worlddreamer} has the capability to generate high-quality, cinematic landscape videos. The resulting videos exhibit seamless frame-to-frame motion, akin to the smooth camera movements seen in real films. Moreover, these videos adhere meticulously to the constraints imposed by the original image, ensuring a remarkable consistency in frame composition. It generates subsequent frames adhering to the constraints of the initial image, ensuring remarkable frame consistency.

\textbf{Text to Video}
Fig.~\ref{fig:t2v} demonstrates \textit{WorldDreamer}'s remarkable proficiency in generating videos from text across various stylistic paradigms. The produced videos seamlessly align with the input language, where the language serves as a powerful control mechanism for shaping the content, style, and camera motion of the videos. This highlights \textit{WorldDreamer}'s effectiveness in translating textual descriptions into visually faithful video content.

\textbf{Video Inpainting}
As depicted in Fig.~\ref{fig:inpaint}, \textit{WorldDreamer} exhibits an exceptional ability for high-quality video inpainting. By providing a mask outlining the specific area of interest and a text prompt specifying desired modifications, \textit{WorldDreamer} intricately alters the original video, yielding remarkably realistic results in the inpainting process.

\textbf{Video Stylization}
Fig.~\ref{fig:style} shows that \textit{WorldDreamer} excels in delivering high-quality video stylization. By supplying a randomly generated visual token mask and a style prompt indicating desired modifications, WorldDreamer convincingly transforms the original video, achieving a genuinely realistic outcome in the stylization process.

\textbf{Action to Video}
\textit{WorldDreamer} shows the ability to generate videos based on actions in the context of autonomous driving. As shown in Fig.~\ref{fig:ad}, given identical initial frames and different driving actions, \textit{WorldDreamer} can produce distinct future videos corresponding to different driving actions (e.g., controlling the car to make a left-turn or a right-turn).

\section{Conclusion}

In conclusion, \textit{WorldDreamer} marks a notable advancement in world modeling for video generation. Unlike traditional models constrained to specific scenarios, WorldDreamer capture the complexity of general world
dynamic environments. \textit{WorldDreamer} frames world modeling as a visual token prediction challenge, fostering a comprehensive comprehension of general world physics and motions, which significantly enhances the capabilities of video
generation. In experiments, \textit{WorldDreamer} shows exceptional performance across scenarios like natural scenes and driving environments, showcasing its adaptability in tasks such as text-to-video conversion, image-to-video synthesis, and video editing.

\newpage

{\small
\bibliographystyle{ieee_fullname}
\bibliography{PaperForReview}
}
\end{document}